\documentclass[twoside,11pt]{article}

\usepackage{jmlr2e}
\usepackage{microtype}
\usepackage{graphicx}
\usepackage{subfig}
\usepackage{booktabs}
\usepackage{times}
\usepackage{footnote}
\usepackage{amsmath}
\usepackage{bigints}
\usepackage{dsfont}
\usepackage{algorithm}
\usepackage[noend]{algpseudocode}

\algrenewcommand\algorithmicforall{\textbf{foreach}}
\algrenewcommand\algorithmicindent{.8em}
\usepackage{mathtools}
\usepackage{booktabs}

\usepackage{latexsym}
\DeclareMathOperator*{\argmax}{argmax}
\usepackage{url}
\firstpageno{1}

\begin{document}

\title{BOSH: Bayesian Optimization by Sampling Hierarchically}

\author{\name Henry B. Moss \email h.moss@lancaster.ac.uk \\
       \addr STOR-i Centre for Doctoral Training, Lancaster University, UK
       \AND
       \name David S. Leslie \email d.leslie@lancaster.ac.uk \\
       \addr Dept. of Mathematics and Statistics, Lancaster University, UK
       \AND
       \name Paul Rayson \email p.rayson@lancaster.ac.uk \\
       \addr School of Computing and Communications, Lancaster University, UK}
\editor{}

\maketitle

\begin{abstract}
Deployments of Bayesian Optimization (BO) for functions with stochastic evaluations, such as parameter tuning via cross validation and simulation optimization, typically optimize an average of a fixed set of noisy realizations of the objective function. However, disregarding the true objective function in this manner finds a high-precision optimum of the wrong function. To solve this problem, we propose \textit{Bayesian Optimization by Sampling Hierarchically} (BOSH), a novel BO routine pairing a hierarchical Gaussian process with an information-theoretic framework to generate a growing pool of realizations as the optimization progresses.  We demonstrate that BOSH provides more efficient and higher-precision optimization than standard BO across synthetic benchmarks, simulation optimization, reinforcement learning and hyper-parameter tuning tasks. 
\end{abstract}

\section{Introduction}
\label{sec::Intro}
Bayesian Optimization (BO) \citep{mockus2012bayesian} is a well-studied global optimization routine for finding the optimizer $\textbf{x}^*$ of a `smooth' but expensive-to-evaluate function $g$ across a compact domain $\mathcal{X}\subset\mathds{R}^d$. BO is particularly popular for problems where we have access to only noisy evaluations of $g$ and has had many successful applications optimizing high-cost stochastic functions, including fine-tuning machine learning (ML) models \citep{snoek2012practical}, optimizing simulations in operational research \citep{kleijnen2009kriging}, and designing experiments in the physical sciences \citep{frazier2016bayesian}.

For many stochastic optimization tasks, it is commonplace to disregard the original objective function $g$ and instead optimize the average of a collection of $K$ specific realizations $f_s$. Common examples include the $K$ data partitions used to estimate ML model performance through $K$-fold Cross Validation (CV) \citep{kohavi1995study} or considering $K$ fixed initial conditions to create sample average approximations  for simulation optimization and reinforcement learning \citep{kleywegt2002sample}. This small collection of realization indexed by $S=\{s_1,..s_K\}$ is typically randomly initialized, but then fixed for the remainder of the optimization. We henceforth refer to $S$ as an \textbf{evaluation strategy}, with its optimization seeking
$\textbf{x}_S^*=\argmax_{\textbf{x}\in\mathcal{X}}\tilde{g}_S(\textbf{x})$, where $\tilde{g}_S(\textbf{x})=\frac{1}{K}\sum_{i=1}^{K}f_{s_i}(\textbf{x})$.

Evaluations of $\tilde{g}_S(\textbf{x})$ enjoy a substantial reduction in variance compared to a single stochastic evaluation of the true objective function $g(\textbf{x})$. However, there is no guarantee that $\textbf{x}^*_S\approx\textbf{x}^*$, as $\textbf{x}^*_S$ is a function of the randomly selected $S$. In fact, estimators of this form are well-studied in the robust statistics literature \citep{hampel2011robust}, where it is known that the expected suboptimality $\mathds{E}_{S}[g(\textbf{x}^*)-g(\textbf{x}_S^*)]$ is a positive quantity decaying as $O(\frac{1}{K})$. Regardless of the sophistication of our optimization routine, if $K$ is set too low we cannot optimize $g$ to an arbitrary precision level by optimizing $\tilde{g}_S$. In contrast, as each individual evaluation of $\tilde{g}_S$ costs $K$ times that of evaluating $g$, setting $K$ too large wastes computational resources on unnecessarily expensive evaluations. Therefore, as demonstrated for hyper-parameter tuning \citep{Moss2018}, model selection \citep{moss2019fiesta} and  simulation optimization \citep{kim2015guide}, the efficiency and effectiveness of a fixed evaluation strategy crucially depends on the choice of $K$.% taking into account evaluation variability and the desired optimization precision. 

\begin{figure}[t]
\centering
\includegraphics[width=\textwidth]{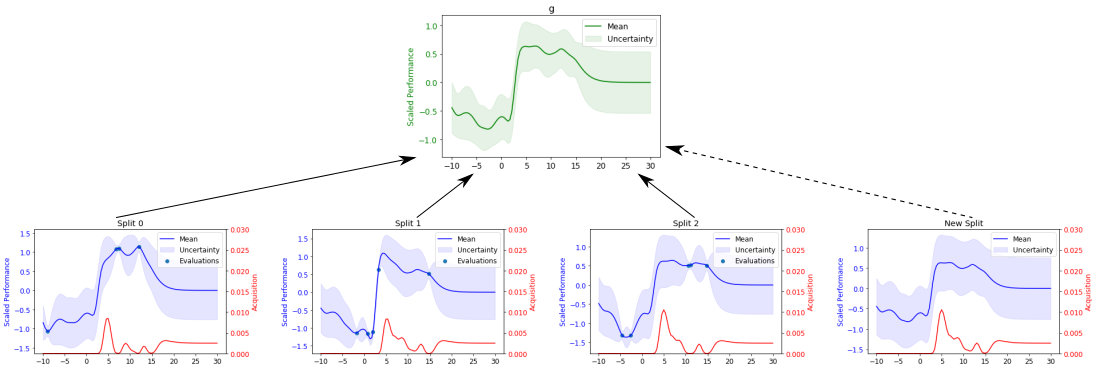}
 \vskip -0.1in
\caption{Tuning an SVM hyper-parameter on  IMDB data using BOSH. Evaluations across three train-test splits (blue) are aggregated to predict the true accuracy (green) and the likely behavior of a new train-test split (right-most panel). The utility of making a new evaluation on each of the considered splits (red) is lower around locations already evaluated on another train-test split. 
\label{SVMplot}}
		\label{Tree}
	\vskip -0.2in
\end{figure}

To avoid the need for fixed evaluation strategies, we  propose \textbf{BOSH} (Bayesian Optimization by Sampling Hierarchically), the  first BO routine that maintains and grows a pool of realizations as the optimization progresses. A \textbf{Hierarchical Gaussian Process} (HGP) \citep{hensman2013hierarchical} is used to quantify uncertainty in our current evaluation strategy by modeling different realizations as separate perturbations of the latent `true' object function (Figure \ref{Tree}). A novel information-theoretic framework then uses the HGP's predictions to balance the utility of making further evaluations in the current pool $\{f_s\}_{s\in S}$ against the benefit of considering a new realization $f_{s^*}$.

\section{Related Work}
\label{related_work}
Using low-cost approximations to speed up optimization is well-studied. Multi-task \citep[MT]{swersky2013multi,poloczek2017multi} and multi-fidelity \citep[MF]{kandasamy2016gaussian,lam2015multifidelity,mcleod2017practical,wu2018continuous,takeno2019multi} BO provides efficient optimization for problems where  low-cost alternative functions hold some relationship with the true objective function. A popular application is hyper-parameter tuning \citep{klein2016fast,kandasamy2017multi,falkner2018bohb}, where dataset size is controlled to provide fast but rough tuning. The closest MT framework to BOSH is FASTCV \citep{swersky2013multi}, which speeds up hyper-parameter tuning under fixed evaluation strategies by choosing to evaluate individual $K$ train-test splits making up $K$-fold CV. However, FASTCV's coregionalisation kernel cannot predict performance on previously unobserved splits or support an adaptive evaluation strategy. To guarantee high-precision optimization, a large choice of $K$ must be chosen \textit{a-priori}, incurring substantial initialization costs and slower optimization. Furthermore, these approaches are unable to recommend batches of points, and it is unclear how to apply existing batch heuristics, for example \citet{gonzalez2016batch}, to MF or MT frameworks.

Parallel work of \citet{pearce2019bayesian} from the operational research literature address a similar problem but in a different way; reducing simulation stochasticity by exploiting common random numbers. Similarly to BOSH, performance is measured according to individual random samples. However, their model is complex and challenging to fit, and their search strategy incurs a computational overhead that grows exponentially with the dimensions of the search space. In contrast, our framework makes principled decisions with a linearly scaling cost and is able to recommend batches.

\section{BOSH}
\label{sec::bosh}
The key difference between BOSH and existing BO is that instead of only modeling $\tilde{g}_S$ for a fixed evaluation strategy $S$, BOSH separately models individual realizations $f_s$. By assuming that each $f_s$ is some perturbation of the true objective function $g$ (see Fig.\ \ref{SVMplot}), we can fit a hierarchical model that learns the correlations between $g$ and each $f_{s}$. Knowledge of this correlation structure provides information about the likely behavior of a yet unobserved realization $f_{s^*}$. Therefore, BOSH can make principled decisions about which realization to use for the next evaluation from the set of candidate realizations $\{f_s\}_{s\in S^*}$, where $S^*=S\cup \{s^*\}$ --- either a realization from  the current evaluation strategy $S$ or generating a new realization $f_{s^*}$ (to be absorbed into $S$ for subsequent optimization steps). This allows BOSH to target $g$ directly, instead of targeting just $\tilde{g}_S$.

Like most BO frameworks, BOSH has two key components: a Gaussian Process (GP) \citep{rasmussen2003gaussian} \textbf{surrogate model} predicting the values of not-yet-evaluated points, and an \textbf{acquisition function} using these predictions to efficiently explore the search space. For BOSH, we require an acquisition function $\alpha$ estimating the utility of evaluating any $\textbf{x}\in\mathcal{X}$ on any realization $f_s$ for $s\in S^*$. After collecting  $n$ (potentially noisy) evaluations, BOSH evaluates locations on realizations that score highly according to the acquisition function, repeating until the optimization budget is exhausted.

\subsection{The BOSH Surrogate Model}
\label{sec::HGP}

\begin{figure}[ht]
\includegraphics[width=0.26\textwidth]{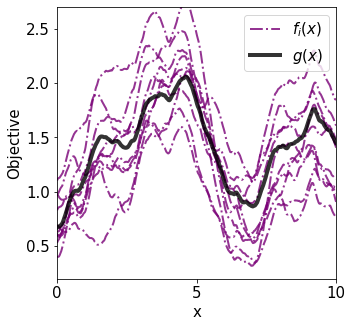}
\includegraphics[width=0.26\textwidth]{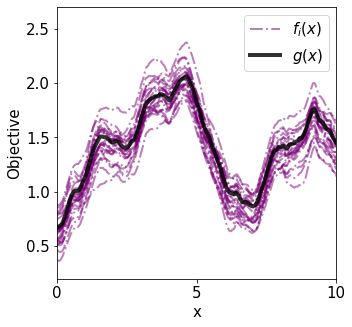}
\includegraphics[width=0.46\textwidth]{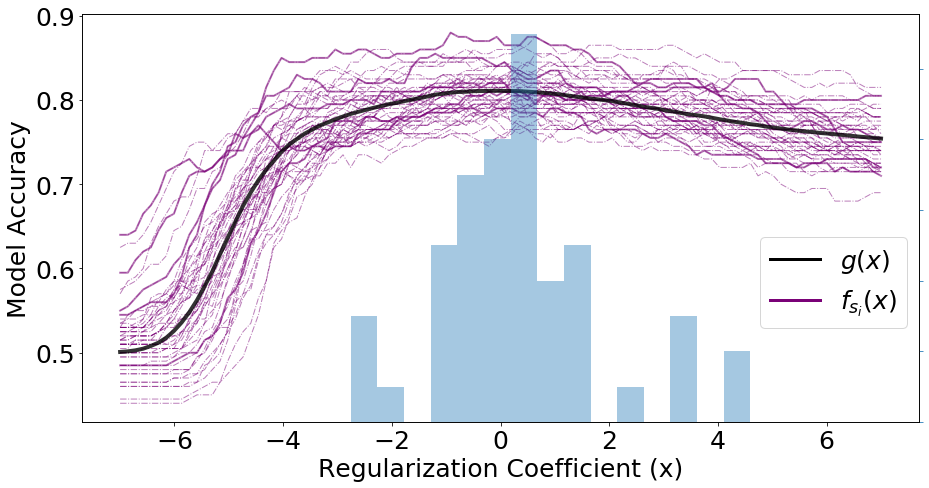}
 \vskip -0.1in
\caption{(left) and (middle) show samples from HGPs with different lower kernels, demonstrating a capacity for modeling real scenarios like the performance estiamted by train-test splits when tuning regularization of a logistic regression sentiment classifier of IMDB movie reviews (right). Purple lines show sampled $f_{s}(x)$ and the true objective $g(x)$ is plotted in black (as calculated on a large independent test set). The histogram of chosen regularization (performance curve maxima) shows many splits choosing highly suboptimal regularization ($-4\%$ accuracy).}
 \vskip -0.1in
\label{HGP}
    \end{figure}

A natural framework for modeling function realizations as perturbations of a true objective function is a Hierarchical Gaussian Process (HGP) \citep{hensman2013hierarchical}, where the true objective function is modeled as a GP with an `upper' kernel $k_g$, and the deviations to all the individual realizations $f_s$ modeled by another GP with a `lower' kernel $k_f$. As is common in BO, we use Mat\'{e}rn 5/2 kernels \citep{matern1960spatial}. The HGP structure is equivalently understood as  each $f_{s}$ being a conditionally independent GPs with shared mean function $g$, i.e.
\begin{align}
y_i = f_{s_i}(\textbf{x}_i)+\epsilon_i \quad \text{for} \quad
f_{s} \sim \mathcal{GP}(g,k_f) \quad \text{where} \quad
g \sim \mathcal{GP}(0,k_g) 
, \nonumber
\end{align}
for $\epsilon_i\stackrel{\rm i.i.d}{\sim}\mathcal{N}(0,\sigma^2)$. This induces a prior covariance structure of 
\begin{align}
Cov(f_s(\textbf{x}),f_{s'}(\textbf{x}')))=k_g(\textbf{x},\textbf{x}')+\mathds{I}_{s=s'}k_f(\textbf{x},\textbf{x}') \quad \text{and} \quad
Cov(f_s(\textbf{x}),g(\textbf{x}'))=k_g(\textbf{x},\textbf{x}'),\nonumber
\end{align}
where $\mathds{I}$ is an indicator function. Samples from this prior are provided in Fig.\ \ref{HGP}. 
Crucially, given observations $D_n=\{(\textbf{x}_i,s_i,y_i)\}_{i=1}^n$, the HGP provides a bi-variate Gaussian joint distribution for $(y_s(\textbf{x}),g(\textbf{x}))\,|\,D_n$,  the quantities required to evaluate our acquisition function (see below). Prediction cost is equivalent to a standard GP, with the $n^{th}$ BO step dominated by an $O(n^3)$ matrix inversion.

\subsection{The BOSH Acquisition Function}
\label{sec::MUMBO}
We base our acquisition function on the max-value entropy search of \cite{wang2017max}, which seeks to reduce uncertainty in the optimal value $g^*=g(\textbf{x}^*)$. As is common in the BO literature \citep{hennig2012entropy,hernandez2014predictive}, we measure uncertainty in terms of the differential entropy of  our current belief about the maximum value, given by $H(g^*)=-\mathds{E}_{g\sim p_{g^*}}\left(\log p_{g^*}(g)\right)$, where $p_{g*}$ is the probability density function of $g^*|D_n$ according to our HGP. The reduction in entropy of $g^*$ provided by a batch of $B$ evaluations $\{y_{s_{j}}(\textbf{x}_j)\}_{j=1}^B$ is measured as their mutual information $I$, defined as
%{
%\setlength{\abovedisplayskip}{0.1in}
%\setlength{\belowdisplayskip}{0.1in}
%\setlength{\abovedisplayshortskip}{0in}
%\setlength{\belowdisplayshortskip}{0in}
\begin{align}
I(g^*;\{y_{s_{j}}(\textbf{x}_j)\}_{j=1}^B|D_n)\coloneqq H(\{y_{s_{j}}(\textbf{x}_j)\}_{j=1}^B|D_n)- \mathds{E}_{g^*|D_n}\left[H(\{y_{s_{j}}(\textbf{x}_j)\}_{j=1}^B|g^*,D_n)\right].\label{acq}
\end{align}%}
Defining $\textbf{z}_i=(\textbf{x}_i,s_i)$, principled batch BO corresponds to selecting $\{\textbf{z}_i\}_{i=1}^B$ to maximize (\ref{acq}).

Unfortunately,  neither $g^*|D_n$ nor the differential entropy of $y_s(\textbf{x}|g^*,D_n)$ have closed-form expressions. Therefore, to implement information-theoretic BO, the second term of (\ref{acq}) must be approximated. The MUMBO (MUlti-task Max-value Bayesian Optimization) acquisition function \citep{mumbo}  provides one such approximation when $B=1$, requiring only simple single-dimensional numerical integrations regardless of the dimensions of the search space. To extend MUMBO beyond $B>1$, we make an additional approximation  through a well-known information-theoretic inequality --- that the joint differential entropy of a collection of random variables is upper-bounded by the individual entropies. BOSH's acquisition function is the resulting lower bound, expressed in terms of the MUMBO acquisition function as

\begin{align}    \label{BATCH_BOSH_ACQ}
    \alpha_n^{BOSH}(\{\textbf{z}_j\}_{j=1}^B) = \frac{1}{2}\log\left(|\textbf{C}_n(\{\textbf{z}_j\}_{j=1}^B)|\right)+\sum_{j=1}^{B}\alpha^{MUMBO}_n(\textbf{z}_j),
\end{align}
where $\textbf{C}_n$ is the HGP's predictive correlation matrix between each of the $B$ batch elements. The first term of (\ref{BATCH_BOSH_ACQ}) encourages diversity (achieving high values for batches with low posterior correlation) whereas the second term encourages evaluations in areas providing large amounts of information about $g^*\,|\,D_n$. As the $B\times d$-dimensional maximization of (\ref{BATCH_BOSH_ACQ}) posed too great a computational challenge, we greedily construct batches with $B$ separate sequential decisions, each performed with the DIRECT maximizer \citep{jones2009direct}. An extended publication (currently in submission) explores the relationship between (\ref{BATCH_BOSH_ACQ}) and determinantal point processes \citep[e.g.][]{kulesza2012determinantal}.

\section{Experiments}
\label{experiments}

All our experiments show that fixed evaluation strategies can provide either precise or efficient optimization of stochastic objective functions, whereas BOSH achieves both. We compare BOSH against standard BO using two popular acquisition functions: expected improvement (EI) \citep{mockus1978application} and max-value entropy search (MES) \citep{wang2017max}. We also consider FASTCV  \citep{swersky2013multi}, and, for our hyper-parameter tuning tasks, FABOLAS \citep{klein2016fast}. Code built upon the Emukit Python package \citep{emukit2019} is provided, alongside additional experimental details, at \textit{redacted for review}. For a fair reflection of parallel computing resources, evaluations of whole batches (or evaluation strategy) are recorded as a single BO step. We compare the performance of BOSH producing batches of size $B$ against the performance of standard BO using evaluation strategies of $K=B$ fixed realizations as well as $B$ realizations re-sampled at each BO step. We plot the mean suboptimality and one standard error across 100  repetitions.

BOSH's ability to evaluate diverse batches of points in parallel instead of single locations across a whole evaluation strategy provides a natural advantage over standard BO, particularly in the early stages of optimization. Therefore, to explicitly disentangle the benefits of BOSH's adaptive evaluation strategy from its ability to recommend batches, we also consider standard BO recommending batches across an evaluation strategy consisting of a single fixed realization. We present the performance of batch BO  choosing $B$ evaluations across a single fixed realization, considering both the popular locally penalized (LP) EI \citep{gonzalez2016batch} as well as our proposed batch approximation applied to a MES acquisition function (instead of MUMBO). Note that FASTCV and FABOLAS do not support batches.  We do not consider simultaneously deploying both batch BO and full evaluation strategies, as this is beyond the resources of most ML researchers.

For GP initialization, we randomly sample one more evaluation than kernel parameters (to guarantee identifiability). For standard BO, this corresponds to d+3 evaluations of the chosen evaluation strategy (i.e  K*(d+3) individual evaluations). For BOSH, rather than using separate lower and upper kernels for our HGP, we found that sharing length-scales between each kernel greatly improved the stability of the HGP and allowed reliable initialization after just d+5 evaluations across an initial pair of realizations. Reliable initialization of FASTCV's K*K correlation matrix entries (of which its performance was sensitive) required at least d+3  evaluations for each of its K realizations. Therefore, as well as providing improved efficiency and precision once optimization begins, BOSH's ability to model only as many individual realizations as required allows significantly lower initialization costs.

\begin{figure}[t]
\centering
\begin{minipage}[b]{0.58\textwidth}
\includegraphics[width=\textwidth]{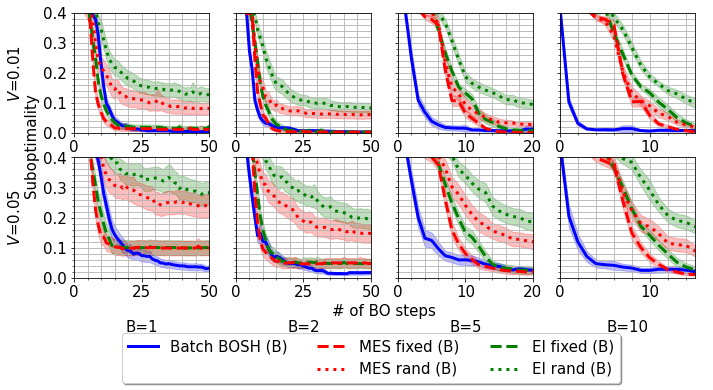}
\parbox{0.95\textwidth}{\caption{ \label{Simulated}Maximizing the upper functions of two HGPs with differing lower kernel variances ($V$) across a range of evaluation strategy sizes ($B$).  BOSH's optimization is more efficient (precise) than BO on large (small) evaluation strategies}}
\end{minipage}%
\begin{minipage}[b]{0.4\textwidth}
\centering
\includegraphics[width=\textwidth]{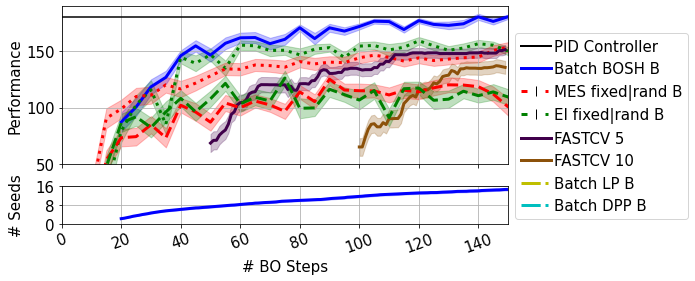}
\includegraphics[width=0.48\textwidth]{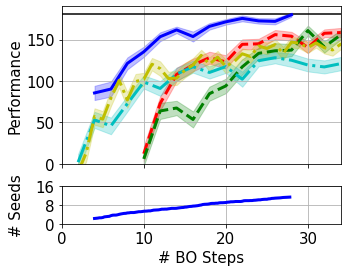}
\hspace{0.1in}
\includegraphics[width=0.41\textwidth]{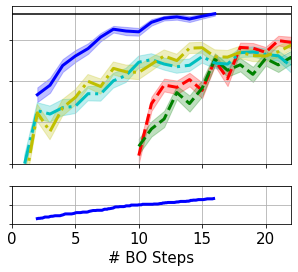}
\parbox{0.95\textwidth}{\caption{\label{LL}Maximizing  Lunar Lander performance with B=1,5,10 (upper, left, right). BOSH adaptively considers up to $15$ realizations and can match the performance of the PID controller.}}  
\end{minipage}
 \vskip -0.35in
\end{figure}

\textbf{Synthetic Objective (d=1)}. First, we simulate data directly from an HGP (Figure \ref{Simulated}) and seek to find the maximum of $g(x)$ (as plotted in Figure \ref{HGP}) by querying only the perturbed curves $f_s$. We consider two  different lower kernels, one with a smaller variance ( $V$) causing low between-realization variability, and another with a larger variance causing high between-realization variability.

\textbf{Reinforcement Learning (d=7)}.
BOSH can fine-tune the $7$ parameters directing a lunar lander module to its landing zone ( provided by OpenAI Gym \textit{https://gym.openai.com/envs/LunarLander-v2/}). A particular configuration is tested by running a single (or $B$) randomly generated scenarios. We seek to outperform OpenAI's hard-coded controller (denoted PID ) according to `true' performance over a set of  $100$ specific initial conditions, using as few simulation runs as possible (Figure \ref{LL}).

\begin{figure}[h]
\begin{minipage}[b]{0.62\textwidth}
\centering
\includegraphics[width=\textwidth]{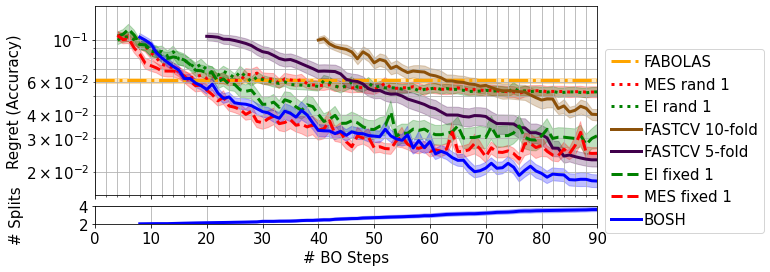}
\vskip -0.05in
\includegraphics[width=0.41\textwidth]{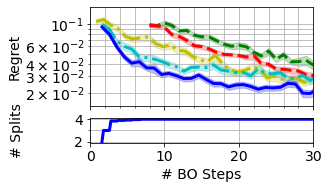}
 \includegraphics[width=0.57\textwidth]{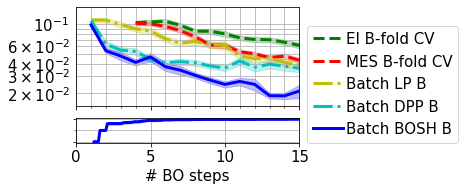}
\parbox{0.95\textwidth}{\caption{ \label{IMDB}Minimizing SVM error for IMDB movie review classification with B=1,5,10 (upper, left, right). BOSH considers up to four realizations to provide higher-precision tuning than standard BO. When parallel resources are available, BOSH provides faster tuning than BO under CV and more precise tuning than batch BO.}}
\end{minipage}%
\begin{minipage}[b]{0.36\textwidth}
\centering
\includegraphics[width=\textwidth]{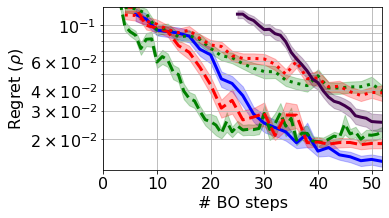}
\vskip -0.05in
\includegraphics[width=\textwidth]{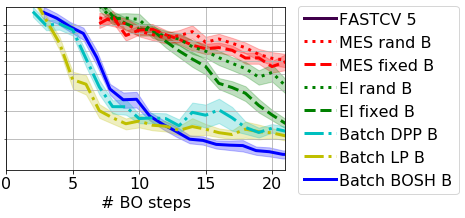}
\parbox{0.95\textwidth}{\caption{\label{fac}Allocating warehouses to cope with demand for B=1,5 (upper, lower). Although standard BO provides fast rough optimization, only BOSH achieves high-precision optimization.}}
\end{minipage}
\vskip -0.35in
\end{figure}

\textbf{Hyper-parameter Tuning (d=2)}.
BOSH can also be used to tune the hyper-parameters of ML models, e.g. the two  hyper-parameters of an SVM classifying IMDB movie review sentiment (Figure \ref{IMDB}). During tuning, BOSH uses a pool of train-test splits and standard BO uses fixed evaluation strategies of single train-test splits or $K$-fold CV. True performance is calculated retrospectively on a large held-out test set. Although finding a reasonable configuration after a very small optimization budget, FABOLAS's reliance on low-fidelity estimates prevents precise optimization.

\textbf{Simulation Optimization (d=4)}.
Our final experiment (Figure \ref{fac}) considers a simulation optimization problem from the set of benchmarks of \textit{http://simopt.org/}. We wish to decide $(x,y)$ locations of two warehousing facilities. Orders arise according to a pre-specified non-homogeneous Poisson process and each order is served by one of the ten trucks belonging to the closest warehouse (or queued if all trucks are busy). The goal is to maximize the proportion $\rho$ of orders delivered within $60$ minutes. Base estimate of $\rho$ comes from simulating demand for a single day according to a single random seed. We can calculate more reliable estimates by simulating demand for $B$ independent days and we retrospectively estimate the true $\rho$ with an expensive but reliable $100$ day simulation.

%\section{Conclusions}
%Optimizing stochastic functions using BO under fixed evaluation strategies does not achieve high precision optimization as it simply results in over-fitting to the evaluation strategy. We instead propose BOSH, a novel framework for deploying BO over an evaluations strategy that grows as the optimization progresses.
\newpage
\def\bibfont{\small}

\end{document}